\documentclass[10pt,twocolumn,letterpaper]{article}

\usepackage{wacv}
\makeatletter
\@namedef{ver@everyshi.sty}{}
\makeatother
\usepackage{times}
\usepackage{epsfig}
\usepackage{graphicx}
\usepackage{amsmath}
\usepackage{amssymb}
\usepackage{booktabs}

\usepackage{subcaption}
\usepackage{tikz}
\usepackage{comment}
\usepackage{amsmath,amssymb} % define this before the line numbering.
\usepackage{color}
\usepackage{xcolor}
\usepackage{cite}
\usepackage[accsupp]{axessibility}  % Improves PDF readability for those with disabilities.
\newcommand{\quotes}[1]{``#1''}

\usepackage{tcolorbox}
\usepackage{wrapfig}
\usepackage{multirow}
\usepackage{colortbl}

\newtcbox{\mybox}[1][]{on line,fontupper=\ttfamily\footnotesize, colframe=black!10,tcbox raise base,shrink tight,#1}

\newcommand\mask{\mybox{{[}MASK{]}}~}
\newcommand\cls{\mybox{{[}CLS{]}}~}
\newcommand\sep{\mybox{{[}SEP{]}}~}
\newcommand\eos{\mybox{{[}END{]}}~}
\newcommand\unk{\mybox{{\textless}UNK{\textgreater}}~}
\newcommand\modelname{VLC-BERT}

 \renewcommand{\paragraph}[1]{
     \newline\noindent\newline\noindent\textbf{#1.} 
 }
\def\wacvPaperID{2067} % Enter the WACV Paper ID here

\wacvalgorithmstrack   % Uncomment this line if you are submitting to the Algorithms Track.
\wacvfinalcopy % *** Uncomment this line for the final submission

\ifwacvfinal
\usepackage[breaklinks=true,bookmarks=false]{hyperref}
\else
\usepackage[pagebackref=true,breaklinks=true,colorlinks,bookmarks=false]{hyperref}
\fi

\begin{document}

%%%%%%%%% TITLE
\title{\modelname{}: Visual Question Answering\\with Contextualized Commonsense Knowledge}

\author{Sahithya Ravi$^{1,2}$\thanks{Denotes equal contribution}~~~Aditya Chinchure$^{1,2\ast}$~~~Leonid Sigal$^{1,2}$ ~~~Renjie Liao$^1$~~~Vered Shwartz$^{1,2}$ \\
$^1$ University of British Columbia~~~$^2$ Vector Institute for AI\\
{\tt\small \{sahiravi, aditya10, lsigal, vshwartz\}@cs.ubc.ca, rjliao@ece.ubc.ca }
% For a paper whose authors are all at the same institution,
% omit the following lines up until the closing ``}''.
% Additional authors and addresses can be added with ``\and'',
% just like the second author.
% To save space, use either the email address or home page, not both
% \and
% Second Author\\
% Institution2\\
% First line of institution2 address\\
% {\tt\small secondauthor@i2.org}
}

\maketitle
\thispagestyle{empty}

% %%%%%%%%% ABSTRACT
% \begin{abstract}
%   The ABSTRACT is to be in fully-justified italicized text, at the top
%   of the left-hand column, below the author and affiliation
%   information. Use the word ``Abstract'' as the title, in 12-point
%   Times, boldface type, centered relative to the column, initially
%   capitalized. The abstract is to be in 10-point, single-spaced type.
%   Leave two blank lines after the Abstract, then begin the main text.
%   Look at previous WACV abstracts to get a feel for style and length.
% \end{abstract}

\label{sec:abstract}
\begin{abstract}
%  There has been a growing interest in solving Visual Question Answering (VQA) tasks that require the model to reason beyond the content present in the image. In this work, we focus on questions that require commonsense reasoning. In contrast to previous methods which retrieve knowledge from static knowledge bases, we generate contextualized knowledge dynamically using Commonsense Transformer (COMET), a knowledge model trained on human-curated knowledge bases. Moreover, we investigate the addition of answers generated from GPT-3 using a few-shot approach. We propose a method to generate, select, and encode external commonsense knowledge alongside visual and textual cues in a new pre-trained Vision-Language-Commonsense transformer model, VLC-BERT. Through our evaluation on the knowledge-intensive OK-VQA dataset and the large-scale VQA 2.0 dataset, we show VLC-BERT is able to learn from explicit commonsense knowledge and implicit knowledge obtained through pre-training. VLC-BERT outperforms several existing models on OK-VQA, achieving state-of-the-art performance.
 
There has been a growing interest in solving Visual Question Answering (VQA) tasks that require the model to reason beyond the content present in the image. In this work, we focus on questions that require commonsense reasoning. In contrast to previous methods which inject knowledge from static knowledge bases, we investigate the incorporation of contextualized knowledge using Commonsense Transformer (COMET), an existing knowledge model trained on human-curated knowledge bases.  We propose a method to generate, select, and encode external commonsense knowledge alongside visual and textual cues in a new pre-trained Vision-Language-Commonsense transformer model, VLC-BERT. Through our evaluation on the knowledge-intensive OK-VQA and A-OKVQA datasets, we show that VLC-BERT is capable of outperforming existing models that utilize static knowledge bases. Furthermore, through a detailed analysis, we explain which questions benefit, and which don’t, from contextualized commonsense knowledge from COMET. Code: \href{https://github.com/aditya10/VLC-BERT}{https://github.com/aditya10/VLC-BERT}
\end{abstract}
%highlight that explicit commonsense knowledge is impactful in producing better answers on both datasets. 

% Our model produces new state-of-the-art results in our experiments with OK-VQA, a challenging knowledge-based VQA dataset.

% Though large-scale pre-trained language models capture some of this knowledge implicitly, scaling up models does not necessarily endow them with commonsense knowledge. 
 
   % hierarchical?

\section{Introduction}
\label{sec:intro}
% First paragraph: VQA task definition
Recent progress in multimodal vision-language learning has been fueled by large-scale annotated datasets for Visual Question Answering (VQA) \cite{VQA, visdial, balanced_vqa_v2, rephrasings, zellers2019vcr}, in which models are presented with questions about an image. To answer questions correctly, models are required to perform scene understanding and learn meaningful connections between the two modalities. In recent years, transformer-based vision and language (VL) models \cite{NIPS2017_3f5ee243,devlin2019bert,li2019visualbert}, pre-trained on large-scale multimodal corpora, have reached impressive accuracies on standard VQA datasets.
% For example, in Figure~\ref{fig:vqa_example}, to answer questions about the image, the model needs to understand that the image depicts a plate with meat, potatoes and bread \adi{May not be necessary to include this example sentence}. 
% Second paragraph: Motivation for KBVQA and gaps in current methods

VQA often necessitates not only visual comprehension of the scene depicted by the image (\eg, ``A plate with meat, potatoes and bread'') but also making inferences about plausible stories behind the image (\eg, ``The plate is likely found at a restaurant''). Humans make such inferences based on prior experience and commonsense knowledge (\eg, ``This is likely a lunch or dinner at a restaurant, people may be enjoying themselves...''). Most existing methods rely on world knowledge implicitly encoded by language models, which often lacks in both accuracy and coverage \cite{rogers-etal-2020-primer}. This is primarily due to the fact that commonsense knowledge is extremely broad, and frequently assumed. Commonsense knowledge learned from text suffers from reporting bias \cite{reportingbias}: over-representation of exceptional facts (\eg, \quotes{people die in accidents}) in text corpora, at the expense of rarely discussed trivial facts known to everyone (\eg, \quotes{people eat}). 
\begin{figure}[t!]
  \centering
  \includegraphics[width=0.2\textwidth]{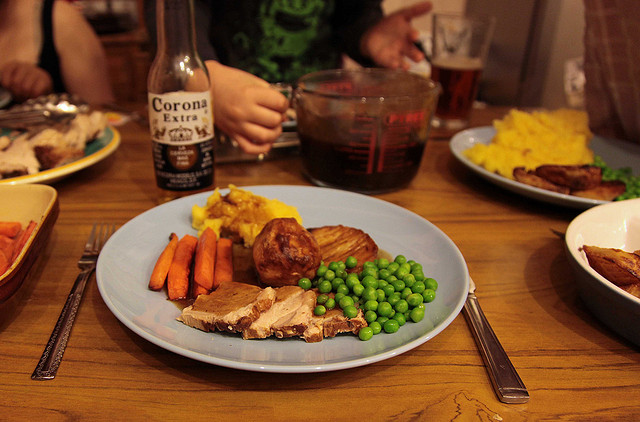}
  \caption{OK-VQA\cite{okvqa}: Where might one buy this?} 
  \label{fig:vqa_example}
\end{figure}

Several visual question answering benchmarks were proposed, in which the questions require either factual \cite{wang2017fvqa, okvqa} or commonsense knowledge \cite{zellers2019vcr, AOKVQA} beyond the visual scene comprehension. This prompted the development of neurosymbolic methods combining transformer-based representations with knowledge bases (KBs) \cite{Gardres2020ConceptBertCR, marino2020krisp, wu2022multi}.
However, retrieving relevant facts directly from a KB is challenging due to lack of coverage, and because KB facts are only appropriate in certain contexts.

In this work, we propose \modelname{} (Vision-Language-Commonsense BERT), a model designed to incorporate contextualized commonsense knowledge into a Vision-Language transformer built on VL-BERT\cite{su2020vlbert}. As an alternative to the retrieval paradigm often used in knowledge-based VQA, our model generates contextualized commonsense inferences on the question phrase combined with image object tags using COMET \cite{bosselut2019comet,Hwang2021COMETATOMIC2O}, a language model trained on commonsense knowledge graphs. We augment sentence transformers\cite{reimers2019sentencebert} to rank, filter and embed the commonsense inferences. We incorporate the filtered inferences into \modelname{} using an attention-driven fusion mechanism that learns to focus on the most important inferences for each question. Commonsense knowledge may not be necessary for answering every question, as some questions are either purely visual, factual, or straight-forward. To eliminate injecting noisy knowledge in such cases, we employ weak supervision to help us discriminate between situations when commonsense knowledge may or may not be valuable.

Our evaluations on the challenging OK-VQA \cite{okvqa} and A-OKVQA \cite{AOKVQA} datasets confirm that leveraging commonsense is consistently useful for knowledge-intensive visual question answering tasks. We analyze the successful predictions and show how the commonsense inferences help answering difficult questions.  
\begin{figure*}%
\centering
\begin{subfigure}{0.6\columnwidth}
\includegraphics[width=\columnwidth]{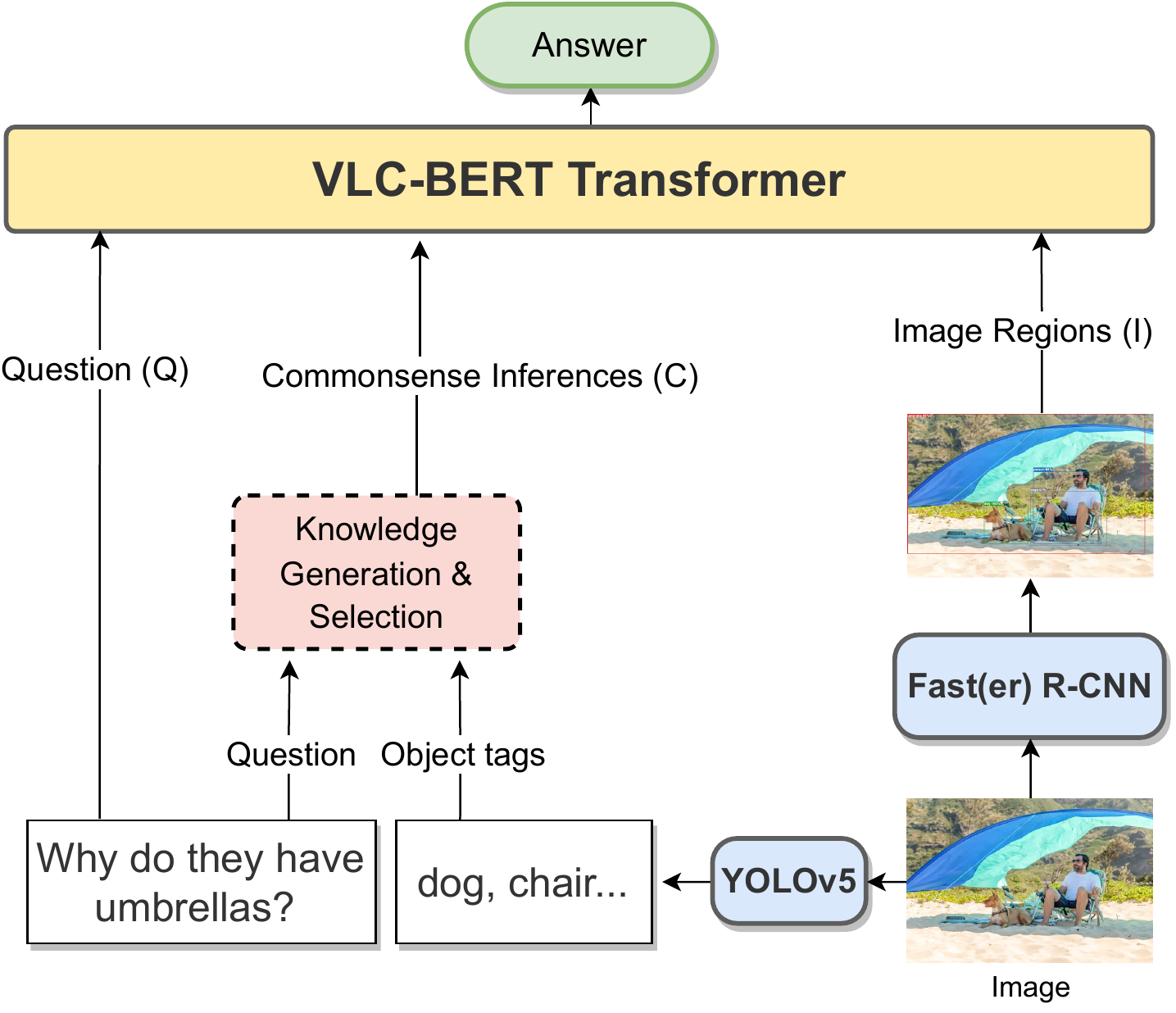}
\caption{Overall architecture}%
\label{fig:architecture}%
\end{subfigure}
\hspace{30pt}%
\begin{subfigure}{1.3\columnwidth}
\includegraphics[width=\columnwidth]{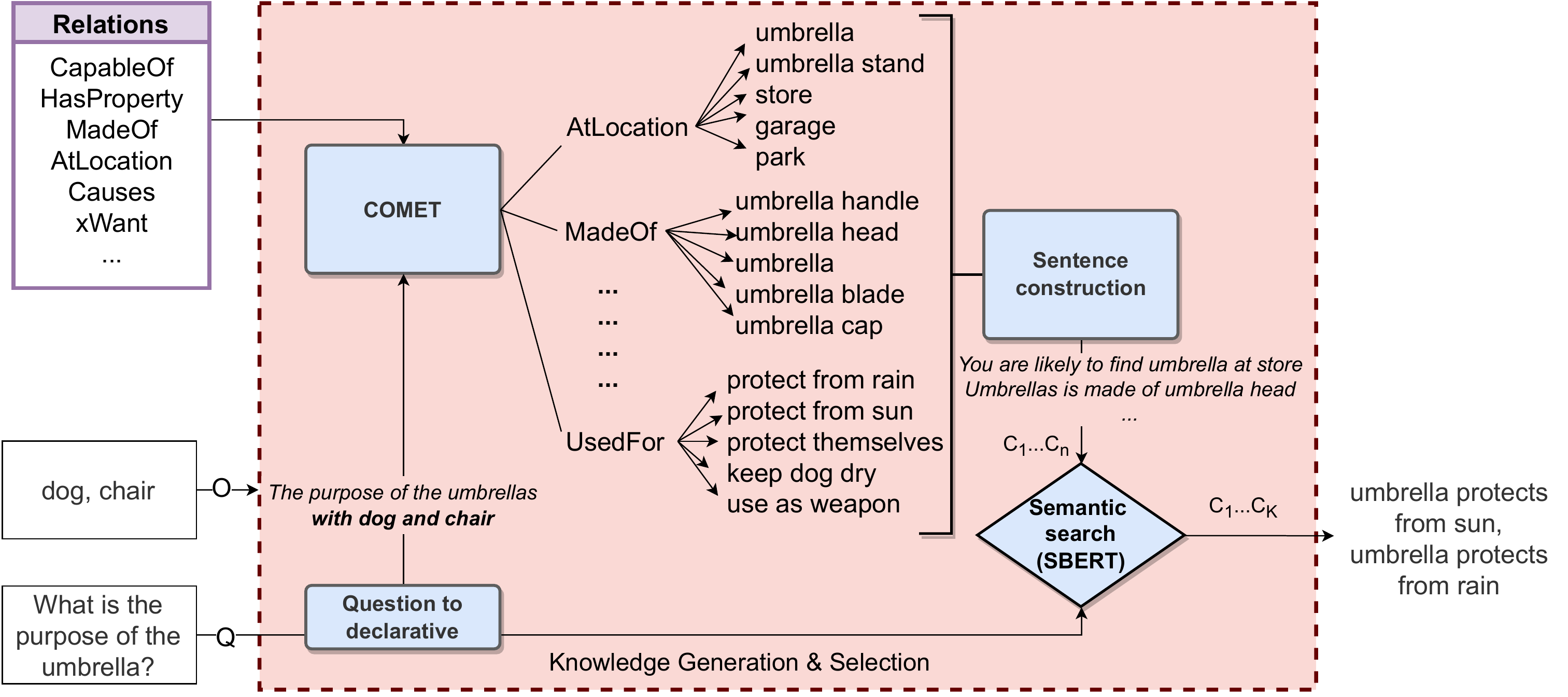}%
  \caption{Knowledge generation and selection}%
  \label{fig:kg}%
\end{subfigure}
\centering
\caption{{\bf Architecture of VLC-BERT}: Given an image, \modelname{} generates commonsense inferences for the question-object phrase using COMET. These inferences are relevance ranked, and top ones are selected and fed along with image regions into a VL-Transformer in order to produce an answer. We utilize semantic similarity between $Q$ and $C$ to select the final $K$ inferences that go into VLC-BERT.}%
\end{figure*}

\section{Related Work}
\label{sec:related_work}
\subsection{Vision-Language Transformer Models}
\label{sec:related_work:vqa}
Pre-trained Vision-Language models based on BERT \cite{devlin2019bert} have shown impressive performances on downstream multimodal tasks such as Visual Question Answering. ViLBERT \cite{Lu_2020_CVPR} and LXMERT \cite{DBLP:journals/corr/abs-1908-07490} use a two-stream architecture to first encode language and vision modalities independently, and then apply a cross-modality encoder to align textual and visual tokens. VL-BERT \cite{su2020vlbert}, OSCAR \cite{li2020oscar} {and OSCAR+ \cite{Zhang_2021_CVPR}} use a single-stream architecture to directly learn inter-modality interactions. Large-scale pre-training is commonly done using the Conceptual Captions \cite{sharma2018conceptual} dataset, with objectives that are designed to encourage interaction between modalities, such as predicting masked tokens or image regions \cite{su2020vlbert, li2020oscar, DBLP:journals/corr/abs-1908-07490, Lu_2020_CVPR}, and using contrastive loss between modalities \cite{li2020oscar}. As a result, such models inherently capture some commonsense knowledge through their pre-training regime. While these models perform impressively on downstream tasks such as VQA\cite{VQA}, they typically perform worse on questions requiring reasoning about knowledge beyond the image content or involving multiple reasoning hops. In our work, we introduce \modelname{}, a multimodal transformer model based on VL-BERT that explicitly incorporates external knowledge to alleviate this issue.
% , such as in OK-VQA \cite{okvqa}, A-OKVQA \cite{AOKVQA} and VCR \cite{zellers2019vcr}.  
\subsection{Knowledge-based Visual Question Answering}
\label{sec:related_work:knowledge_vqa}
In recent years, several VQA datasets were designed specifically to require reasoning about external knowledge beyond the image, whether using factual and web information (FVQA \cite{wang2017fvqa}, WebQA \cite{Chang_2022_CVPR}, a provided text passage (VLQA \cite{sampat-etal-2020-visuo}), commonsense-driven reasoning (VCR \cite{zellers2019vcr}), or external commonsense knowledge (OK-VQA \cite{okvqa}, A-OKVQA\cite{AOKVQA}).
This motivated a line of work on knowledge-enhanced VL transformer models. External knowledge is typically retrieved from a structured knowledge base like ConceptNet \cite{speer2018conceptnet}, in the form of a subgraph, and integrated into the VL transformer as an additional input \cite{Gardres2020ConceptBertCR,Li2020BoostingVQ,marino2020krisp,wu2022multi}. Alternative sources of knowledge include image captions \cite{imagecaptioning}, Google Search results \cite{luo2021weaklysupervised}, and textual and visual knowledge from Wikipedia, and Google Images \cite{wu2022multi}. In contrast to most of the preceding work, PICa \cite{yang2021empirical} and  Knowledge Augmented Transformer (KAT) \cite{gui2021kat} attempt to use GPT-3 \cite{gpt3} in a few-shot setting on the VQA task, by building prompts containing the caption and object tags generated using the image, followed by the question statement, asking the model to produce an answer.  In our proposed model, we focus on a specific subset of the knowledge-intensive datasets that require commonsense knowledge.  % We believe that 
Our approach, that uses COMET \cite{Hwang2021COMETATOMIC2O}, for incorporating commonsense knowledge is distinctly different, far simpler and more cost-effective. 
% and far simpler, as we use COMET \cite{Hwang2021COMETATOMIC2O}, and more cost-effective, considering the GPT-3 API usage in PICA or KAT. 
% We also evaluate our model on the newer and larger A-OKVQA dataset.\vered{consider removing the part about PICa and GPT-3 if we need to save space (and we don't refer to this later).}
\subsection{Knowledge incorporation in NLP}
\label{sec:related_work:knowledge_nlp}
Structured large-scale knowledge bases (KBs) like ConceptNet \cite{speer2018conceptnet} and ATOMIC \cite{sap2019atomic} are widely used in NLP tasks to provide additional commonsense knowledge to models. ConceptNet contains 3.4M assertions focusing on concept and entity relations (such as RelatedTo, Synonym, IsA, MadeOf). ATOMIC contains 1.33M triplets focusing on event-centric social commonsense about causes, effects, mental states of the event participants. 
Several approaches were proposed for incorporating symbolic knowledge from these KBs into downstream NLP tasks such as encoding subgraphs of relevant knowledge  \cite{kagnet-emnlp19,Gardres2020ConceptBertCR} and pre-training on commonsense knowledge bases or tasks \cite{zhong2019improving}.
Despite the performance improvements, incorporating knowledge directly from KBs suffers from two limitations: lack of coverage and lack of consideration for context. Commonsense Transformer, COMET \cite{Hwang2021COMETATOMIC2O}, attempts to alleviate these issues by fine-tuning pre-trained language models on KBs. COMET can generate inferences for the various KB relations dynamically for new inputs. It has been successfully used for generating knowledge in language tasks \cite{majumder-etal-2020-like, tian2021hypogen, Chakrabarty2021ItsNR, shwartz-etal-2020-unsupervised}. Inspired by the success of these models, we chose to use COMET \cite{Hwang2021COMETATOMIC2O} to generate relevant contextual expansions rather than directly retrieving knowledge from KBs. To the best of our knowledge, we are the first to incorporate commonsense knowledge using COMET in VQA tasks.
Newer COMET variants\cite{visualcomet, West2022SymbolicKD} are less applicable to OK-VQA and A-OKVQA as they focus more on event commonsense than entities.

\section{Method}
\label{sec:method}
We briefly outline the overall architecture of our model and then delve deeper into its individual components. Figure~\ref{fig:architecture} illustrates the \modelname{} pipeline. Given an image with corresponding image regions $I$ precomputed using Fast RCNN \cite{girshick2015fast} and a question $Q$ related to the image, we generate commonsense inferences $C$ on the events and entities in the question phrase and two object tags $O$, and select the set of commonsense inferences which is the most useful for answering the question, $C$ = \{$C_1, C_2, ..., C_k$\} (\S\ref{sec:method:comet}). Finally, we embed $Q$, $I$ and $C$, as input to \modelname{} and train it to predict an answer $A$ to $Q$ (\S\ref{sec:method:vlc-bert}). 

\subsection{Structured knowledge generation and selection}
\label{sec:method:comet}

\subsubsection{Knowledge Generation}
\label{sec:method:comet:knowledge_generation}
To generate commonsense knowledge, we employ the most recent version of COMET \cite{Hwang2021COMETATOMIC2O} initialized using BART \cite{lewis2019bart} in a zero-shot setting. COMET is trained to complete 50 relation types from both ConceptNet \cite{speer2018conceptnet} (such as AtLocation, Madeof) and ATOMIC \cite{sap2019atomic} (such as xNeed, xWants), thus capturing concept as well as event oriented knowledge. We generate inferences based on 30 relation types most relevant to our work and supported by COMET.\footnote{We include the full list of relation types in the supplementary material.}Consider the example shown in Figure~\ref{fig:kg}. For the given question, \textit{\quotes{What is the purpose of the umbrella?}} we first process each question using AllenNLP's constituency parser \cite{Joshi2018ExtendingAP} and convert it into a declarative sentence, since COMET was mainly trained on declarative sentences. In the example shown, \textit{\quotes{What is the purpose of the umbrella?}} is rephrased as \textit{\quotes{The purpose of the umbrellas is}}. We then adopt a state-of-the-art object detection model, YOLOv5 \cite{yolov5}, to translate the corresponding image into object tags that COMET can understand. We select the top two most confident object tags and combine it with the question phrase to obtain a question-object(QO) phrase, \textit{\quotes{The purpose of the umbrella is, with dog and chair}}. We restrict the number of the object tags used in COMET's input to two because the addition of multiple tags make the inferences more conflated and noisy. In this manner, we can obtain inferences that can provide additional knowledge about both the visual and language inputs to \modelname{}.

We use beam search to decode the top 5 inferences for each relation type, ranked according to the model's confidence. Overall, we get $30 \times 5 = 150$ inferences for each input phrase. Finally, we convert each inference to a sentence in natural language using relation-specific templates as defined in \cite{davison-etal-2019-commonsense}. In the shown example, the assertion $<umbrella, \textit{Located At}, store>$ is expressed as ``You are likely to find umbrella at store''. In order to remove redundant sentences of the same relation type, we measure the lexical overlap by measuring the percentage of common words between two given sentences. We exclude the sentences which have more than 70\% overlap with previously constructed sentences of the same relation.  
% \paragraph{Augmented-COMET} In this work, we create a version of COMET that is tailored and augmented for our task. The training data for this version is derived from the human-annotated rationales in the training set of A-OKVQA\cite{AOKVQA}. We  extract subject $S$, relation $R$ and target $T$ triplets from the rationale sentences using an Open Information Extractor (OpenIE) \cite{StanfordOpenIEWrapper}. We filter the triplets by removing duplicates and triplets with short or invalid subjects (\eg stopwords, named entities). Then, the relations obtained are mapped to corresponding relations in COMET by using simple template matching (\eg \quotes{can} is mapped to {CapableOf}) \footnote{We will provide the mapping templates and relation distributions in supplementary material}. Our final training set contains *** triplets. We train COMET for one epoch by providing triplets ($S$,$R$,$T$) in this form: \texttt{$S^{(i)}$ $R^{(i)}$ [GEN] $T^{(i)}$}. \vered{Also, consider adding a short example if space permits.}
\subsubsection{Knowledge Selection}
\label{sec:method:comet:knowledge_selection}
Due to the high cost of computation, and the noise associated with feeding such a large number of text tokens, feeding up to 150 COMET inferences into the VL Transformer model is impractical. In order to rank and select the inferences, we employ semantic search based on sentence transformers (SBERT) \cite{reimers2019sentencebert}, which are pre-trained on tasks that retrieve candidate answers to a search query. In this method, the question and the inferences are embedded into the same vector space using SBERT \cite{reimers2019sentencebert} and cosine similarity between the question and the inference embeddings is used to rank the inferences. We prune the set of inference sentences $C$ by picking $K = 5$ inferences which are expected to be the most useful for answering the question $Q$.
\paragraph{Augmented-SBERT} We augment the SBERT used for semantic search by starting with a pre-trained SBERT model and continuing to train it for 2 epochs on question-inference instances from the \emph{training set} of our datasets.  To achieve this, we label the inferences for each question with similarity scores based on the proportion of overlap with the human-annotated answers. Since SBERT is trained on corpora that are distinct from our task, the augmentation ensures that the model understands the nature of query-inference pairings in our tasks. The augmented SBERT especially helps with narrowing down the right relations to the question. For instance, the question in shown in Figure~\ref{fig:kg} benefits most from the relations that talk about what the umbrella (\textit{UsedFor}) is used for or capable of (\textit{CapableOf}.) %\vered{consider adding a short example if space permits.}
\subsection{\modelname{}}
\label{sec:method:vlc-bert}

\begin{figure*}[t]
\centering
    \includegraphics[width=0.6\textwidth, scale=0.1]{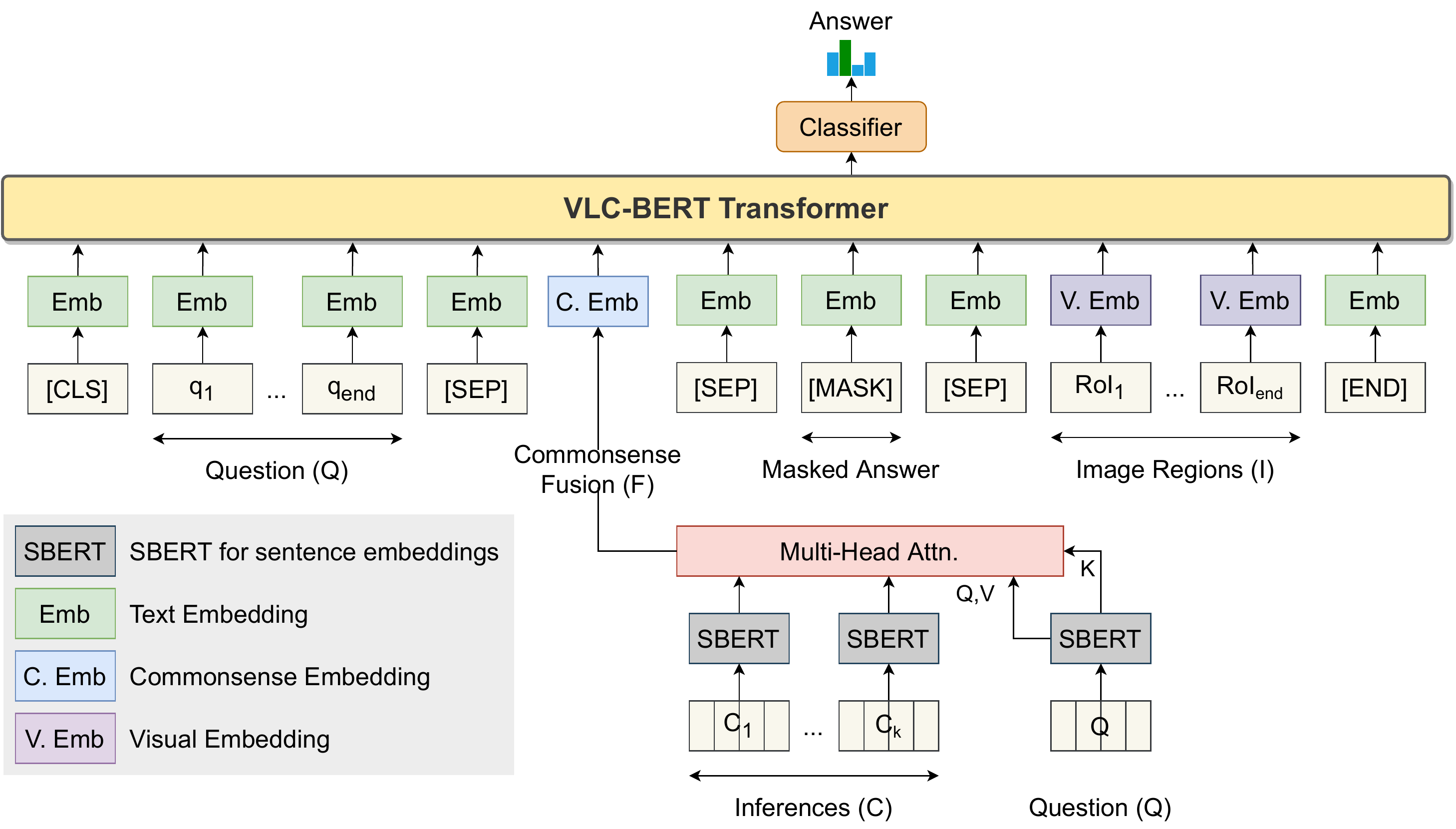}
    \caption{\textbf{\modelname{} Transformer} is a single-stream Transformer that can attend across language, vision, and commonsense representations. We use the $\operatorname{MHA}$ block to fuse commonsense inferences into a useful commonsense representation.}
    \label{fig:vlc}
\end{figure*}

We use a single-stream multimodal transformer encoder, VL-BERT \cite{su2020vlbert}, as the basis of \modelname{}. VL-BERT is pre-trained on large-scale vision-language and language-only datasets with a goal of aligning the visual and linguistic features and building robust multimodal representations for downstream tasks. It is trained on the vision-language Conceptual Captions dataset \cite{sharma2018conceptual}, to predict regions-of-interests (RoIs) from language cues, and on the language-only BookCorpus \cite{bookcorpus} and English Wikipedia corpora, with a masked language modeling objective. 
% We use a single-stream multimodal transformer encoder, VL-BERT \cite{su2020vlbert}, as the basis of \modelname{}. VL-BERT is pre-trained on a large vision-language dataset, Conceptual Captions \cite{sharma2018conceptual}, as well as two large language-only datasets, BookCorpus \cite{bookcorpus} and English Wikipedia. With a goal to align visual and linguistic tokens and build robust multimodal representations for downstream tasks, the model is pre-trained with the objectives of Masked Language Modelling (with visual cues when training on the Conceptual Captions dataset) and RoI classification using language cues. 
%\todo[inline]{cite these datasets. Also say a few words about the pre-training objectives.}
Figure~\ref{fig:vlc} shows the \modelname{} Transformer architecture. In the following paragraphs, we share how the input sequence is constructed and how the predicted answer is selected.

\subsubsection{Inputs} 

Like VL-BERT, \modelname{} accepts word token embeddings for language inputs and RoI token embeddings from the image for vision inputs. The architecture of \modelname{} Transformer is shown in Figure~\ref{fig:vlc}. We use the \cls in the beginning of the sequence, \eos to mark the end of the sequence, and the separator token \sep between different inputs. We feed the question $Q$ as a sequence of word tokens and the image regions $I$ as sequences of RoIs. A \mask token is used to represent the unknown answer. In addition, we introduce a commonsense fusion token, $F$, to the input sequence, to incorporate our commonsense inferences.

A straightforward way to leverage the commonsense inferences $C$ = \{$C_1, C_2, ..., C_k$\} is to embed each word token in every inference sentence as an input token. However, this would lead to a very long input sequence, where the majority of inputs consist of inferences, thus potentially drawing the model's attention away from the other inputs. To overcome the challenge, we summarize the information contained in each inference sentence $C_i$ into a single token representation $\vec{C_i}$, by embedding the inference using SBERT \cite{reimers2019sentencebert}:
\begin{align}
 \vec{C_i} = \operatorname{SBERT}(C_i)
\end{align}

Next, in order to obtain a fused representation of the $k$ commonsense inferences,  we attend to the corresponding SBERT embeddings, $[\vec{C_i}...\vec{C_k}]$ against the SBERT embedding of the question, $\vec{Q}=\operatorname{SBERT}(Q)$. The intuition behind this approach is that the model learns to assign a higher score to the most important inference to the question. The key ($K_{A}$), query ($Q_{A}$) and value ($V_{A}$) are assigned as shown below, 
\begin{align}
    K_{A} & = \vec{Q} \\
    Q_{A}, V_{A} & = \operatorname{append}([\vec{C_i}...\vec{C_k}], \vec{Q}) \\
    \vec{F} & = \operatorname{MHA}(K_{A}, Q_{A}, V_{A}) 
\end{align}
% \begin{align}
%     \vec{C_i} = \operatorname{SBERT}(C_i) \\
%     \vec{Q} = \operatorname{SBERT}(Q) \\
%     K = \vec{Q} \\
%     Q, V = \operatorname{append}([\vec{C_i}...\vec{C_K}], \vec{Q}) \\
%     \vec{F} = \operatorname{MHA}(K, Q, V) 
% \end{align}
\noindent where $\operatorname{MHA}$ is the standard multi-head attention \cite{NIPS2017_3f5ee243}, that delivers a \emph{single vector} incorporating all relevant commonsense knowledge required to answer the question. Note that we append the question embedding $\vec{Q}$ to list of commonsense inference embeddings for $Q$ and $V$ because there may be cases where none of the inferences are useful to answer the question. In such a case, the model may choose to ignore the inferences by attending to the question embedding $\vec{Q}$ instead.
\paragraph{Weak Supervision} In order to train the $\operatorname{MHA}$ block effectively, we employ weak supervision on the attention weights. For a small subset of the questions in the training set, we obtain label attention weights by following these steps: (1) we initialize a vector $\hat{A}$ of length $k+1$ where all values are $0.05$, (2) for each $C_i$, if $C_i$ contains a word in the ground-truth answer list, then we set the $\hat{A}_i$ to $0.8$, (3) if none of the $C$ inferences contain answer words, we assign a weight of $0.8$ to $\hat{A}_{k+1}$ so that the question has the largest weight, and (4) we normalize $\hat{A}$ so that its values sum up to $1$. We then apply cross-entropy loss between the predicted attention weights from $\operatorname{MHA}$ and our label attention weights $\hat{A}$, and sum this with the answer prediction loss.

Finally, a positional encoding is added to all input tokens following the method described in VL-BERT. In addition, a different segment type encoding is applied to the four segments in the input sequence: the question segment, the commonsense segment, the masked answer segment, and the image region segment.

\subsubsection{Answer Selection} 

We use the encoded \mask token to represent the answer, thereby making VQA a masked language modelling task with visual cues. To predict the final answer, we apply a classifier over the entire answer vocabulary, as done in VL-BERT. During training, we follow VL-BERT and use a cross-entropy loss over picking the correct answer from an answer vocabulary.

\section{Datasets}
\label{sec:datasets}
We perform experiments on the OK-VQA \cite{okvqa} and  A-OKVQA\cite{AOKVQA} datasets. In order to utilize the existing VL-BERT model effectively, we pre-train \modelname{} on the larger VQA 2.0 \cite{balanced_vqa_v2}.
\paragraph{\bf OK-VQA}
In the Outside-Knowledge VQA dataset, questions require external knowledge in addition to the information in the images. The dataset is composed of 14,031 images and 14,055 questions, and the crowsourced questions are divided into ten knowledge categories: Vehicles and Transportation; Brands, Companies and Products; Objects, Materials and Clothing; Sports and Recreation; Cooking and Food; Geography, History,
Language and Culture; People and Everyday Life, Plants and Animals; Science and Technology; and Weather and Climate. OK-VQA only contains open-ended questions with five human-provided answers. Since OK-VQA does not have a validation set, we dedicate 1,000 of the 9,009 training questions for validation.
\paragraph{\bf A-OKVQA}
A-OKVQA\cite{AOKVQA} is the augmented successor to OK-VQA and consists of 25K questions that require a combination of commonsense, visual, and physical knowledge.

In contrast to other knowledge-based visual question answering datasets, the questions in A-OKVQA are conceptually diverse, involving knowledge that is not contained in the image, and cannot be resolved by a simple knowledge base query. A-OKVQA is split into training, validation, and test sets based on images used from the COCO 2017 \cite{lin2015microsoft} dataset. Moreover, all questions in the dataset have human annotated direct answers as well as multiple-choice options, but we focus on the direct answers. The A-OKVQA test set is blind, requiring us to submit to the leaderboard to obtain a test accuracy.
\paragraph{\bf VQA 2.0} The Visual Question Answering (v2.0) dataset contains 1.1 million crowdsourced questions about 204,721 images from the COCO dataset \cite{lin2015microsoft}. Each question is annotated with 10 ground truth answers obtained using Amazon Mechanical Turk. A majority of the questions in this dataset do not require external commonsense knowledge.
\setlength{\tabcolsep}{4pt}
\begin{table*}[t]
\begin{center}
\caption{Accuracy of our model against other models for OK-VQA and A-OKVQA datasets. Our model improves upon existing knowledge base based models due to the contextualized commonsense inferences from COMET, which is trained on ConceptNet and ATOMIC. We compare favourably against the highlighted models that utilize external knowledge bases. Note: P.T. stands for Pre-Training.}
\label{table:main}
% \todo[inline]{Cite all the sources for each of the models. @Leon/@Vered/@Renjie do you think we should show all the old models here? It takes up a lot of valuable space. We can choose to only show only newer models on the OK-VQA v1.1 rather than all.}
\begin{tabular}{llccc}
\hline\noalign{\smallskip}
Method & Knowledge Sources & OK-VQA & A-OKVQA & Approx. Params \\
\noalign{\smallskip}
\hline
\noalign{\smallskip}
ViLBERT \cite{AOKVQA} & - & - & 25.85 & 116M  \\
LXMERT \cite{AOKVQA} & - & - & 25.89 & - \\
% ArticleNet (AN) \cite{okvqa} & Wikipedia & 5.28 & - \\
% Q-only \cite{okvqa} & - & 14.93 & - \\
% MLP \cite{okvqa} & - & 20.67 & - \\
% BAN \cite{okvqa} & - & 25.17 & - \\
BAN + AN \cite{okvqa} & Wikipedia & 25.61 & - & - \\
\rowcolor{gray!20} BAN + KG-AUG \cite{Li2020BoostingVQ} & Wikipedia + ConceptNet & 26.71 & - & - \\
% MUTAN \cite{okvqa} & - & 26.41 & - \\
MUTAN + AN \cite{okvqa} & Wikipedia & 27.84 & - & - \\
\rowcolor{gray!20} ConceptBert \cite{Gardres2020ConceptBertCR} & ConceptNet & 33.66 & - & 118M \\
\rowcolor{gray!20} KRISP \cite{marino2020krisp} & Wikipedia + ConceptNet & 32.31 & 27.1 & 116M\\
\rowcolor{gray!20} KRISP \cite{marino2020krisp} & Wikipedia + ConceptNet + VQA P.T. & 38.9 & - & 116M \\
Visual Retriever-Reader \cite{luo2021weaklysupervised} & Google Search & 39.2 & - & - \\
% MAVEx \cite{wu2022multi} & Wikipedia + ConceptNet & 39.45 & - \\
\rowcolor{gray!20} MAVEx \cite{wu2022multi} & Wikipedia + ConceptNet + Google Images & 41.37 & - & - \\
% KAT-explicit \cite{gui-etal-2022-kat} & Wikidata & 44.25 & - \\
GPV2 \cite{gpv2, AOKVQA} & Web Search (Web10k) + COCO P.T. & - & 40.7 & 220M \\
\hline
PICa-Base \cite{yang2021empirical} & GPT-3 & 43.3 & - & 175B\\
PICa-Full \cite{yang2021empirical} & GPT-3 & 48.0 & - & 175B \\
KAT \cite{gui-etal-2022-kat} & Wikidata + GPT-3 & 54.41 & - & 175B\\
\hline
\rowcolor{gray!20} \modelname{} (Ours) & VQA P.T. + COMET & 43.14 & 38.05 & 118M \\
% \modelname{} (Ours) & VQA P.T. + COMET + GPT-3 & \textbf{48.63} & -  \\
\hline
\end{tabular}
\end{center}
\end{table*}
\setlength{\tabcolsep}{1.4pt}
\subsection{Evaluation Metric}
\label{sec:datasets:eval_metric}
Both datasets use the same accuracy-based evaluation metric. Each question has a set of 10 ground truth answers provided by different annotators. Accuracy is calculated as the percentage of predicted answers that were proposed by at least 3 human annotators: $\textrm{acc} = \textrm{min}(\frac{\textrm{ \# humans gave the answer}}{3}, 1)$.\footnote{Following the same evaluation, each of the 5 answers in OK-VQA is used twice}

\section{Implementation Details}
\label{sec:implementation}
%\todo[inline]{Soft evaluation metric during training?}
The implementation of our model builds on VL-BERT \cite{su2020vlbert}. To that end, we follow the fine-tuning steps provided in the official codebase of the VL-BERT model for VQA 2.0, and modify it to support the OK-VQA and A-OKVQA datasets. We maintain the recommended hyperparameter values, and train the \textit{$BERT_{BASE}$} size of the model, with a hidden feature dimension of 768. The model is trained for 20 epochs on the OK-VQA and A-OKVQA datasets. For all models, we use a batch size of 16 and gradient accumulation step size of 4. We train the models presented in the main result thrice and report the average test accuracy on the OK-VQA dataset, and the best (leaderboard) test accuracy on the A-OKVQA dataset.

\noindent \textbf{Answer Vocabulary.}
Due to the large number of unique answers to questions in visual question answering datasets, it is infeasible to use all answers in the answer vocabulary. For the OK-VQA dataset, following KRISP \cite{marino2020krisp}, we build an answer vocabulary of 2,249 answers by selecting all answers in the training set that appear at least 10 times. This answer vocabulary ignores the empty space answer, and includes an \unk answer token. During training, if a ground truth answer is not present in the answer vocabulary, we assign it to the (\unk) token. For the A-OKVQA dataset, we use the answer dictionary that is already provided in the dataset \cite{AOKVQA}.

\noindent \textbf{VQA Pre-Training (VQA P.T)} Following the idea that pre-training is beneficial for Transformer models, we initialize \modelname{} with weights obtained after fine-tuning VL-BERT on the VQA 2.0 dataset for 5 epochs. Note that KRISP \cite{marino2020krisp} benefits from pre-training on the VQA 2.0 dataset, and PICa \cite{yang2021empirical} and KAT \cite{gui-etal-2022-kat} utilize GPT-3, a large-scale pre-trained model, for external commonsense. Furthermore, because OK-VQA and A-OKVQA are significantly smaller than VQA 2.0, this initialization favourably benefits the training process and gives us a stronger baseline to work with.

\section{Evaluation}
\label{sec:eval}
% \setlength{\tabcolsep}{4pt}
% \begin{table}[t]
% \begin{center}
% \caption{Ablation of various components in \modelname{}, evaluated on the A-OKVQA validation set.}
% \label{table:ablation}
% \begin{tabular}{ccccc}
% \hline\noalign{\smallskip}
% VQA P.T. & Aug. SBERT & SBERT & Attn. & Val \\
% \noalign{\smallskip}
% \hline
% \rowcolor{gray!20} \multicolumn{5}{c}{\textbf{VQA Pre-training}} \\
% -- & -- & -- & -- & 36.24 \\
% \checkmark & -- & -- & -- & 43.46 \\
% % \hline
% % \rowcolor{gray!20} \multicolumn{6}{c}{\textbf{Use of Object Tags}} \\
% % \checkmark & -- & \checkmark & \checkmark & \checkmark & \textbf{44.95} \\
% % \checkmark & \checkmark & \checkmark & \checkmark & \checkmark & 44.23 \\
% \hline
% \rowcolor{gray!20} \multicolumn{5}{c}{\textbf{Augmentation of S-BERT}} \\
% \checkmark & -- & \checkmark & \checkmark & 44.10 \\
% \checkmark & \checkmark & \checkmark & \checkmark & \textbf{44.95} \\
% \hline
% \rowcolor{gray!20} \multicolumn{5}{c}{\textbf{Comm. Inference Representation}} \\
% \checkmark &  \checkmark & -- & -- & 43.44 \\
% \checkmark & \checkmark & \checkmark & -- & 43.64 \\
% \checkmark & \checkmark & \checkmark & \checkmark & \textbf{44.95} \\
% \hline
% \end{tabular}
% \end{center}
% \end{table}
% \setlength{\tabcolsep}{1.4pt}

\setlength{\tabcolsep}{4pt}
\begin{table}[t]
\begin{center}
\caption{Ablation of various components in \modelname{}, evaluated on the A-OKVQA validation set. We observe that all the components of our model play a critical role in empirical performance.}
\label{table:ablation}
\begin{tabular}{ccccc}
\hline\noalign{\smallskip}
VQA P.T. & Aug. SBERT & SBERT & Attn. & Val \\
\noalign{\smallskip}
\hline
\rowcolor{gray!20} \multicolumn{5}{c}{\textbf{VQA Pre-training}} \\
-- & -- & -- & -- & 36.24 \\
\checkmark & -- & -- & -- & 43.46 \\
% \hline
% \rowcolor{gray!20} \multicolumn{6}{c}{\textbf{Use of Object Tags}} \\
% \checkmark & -- & \checkmark & \checkmark & \checkmark & \textbf{44.95} \\
% \checkmark & \checkmark & \checkmark & \checkmark & \checkmark & 44.23 \\
\hline
\rowcolor{gray!20} \multicolumn{5}{c}{\textbf{Comm. Inference Representation}} \\
\checkmark &  \checkmark & -- & -- & 43.44 \\
\checkmark & \checkmark & \checkmark & -- & 43.64 \\
\checkmark & \checkmark & \checkmark & \checkmark & \textbf{44.95} \\
\hline
\rowcolor{gray!20} \multicolumn{5}{c}{\textbf{Augmentation of SBERT}} \\
\checkmark & -- & \checkmark & \checkmark & 44.10 \\
\checkmark & \checkmark & \checkmark & \checkmark & \textbf{44.95} \\
\hline
\end{tabular}
\end{center}
\end{table}
\setlength{\tabcolsep}{1.4pt}

In this section, we focus on evaluating \modelname{} on the OK-VQA and A-OKVQA datasets and comparing against existing state-of-the-art models for VQA with external commonsense knowledge. Table \ref{table:main} highlights our performance improvements on the test set for OK-VQA and A-OKVQA against other models. Later in this section, we ablate on the components of our model. 

\subsection{Main Results}
\label{sec:main_results}

Table~\ref{table:main} specifies which knowledge sources each model leverages. In the top section, we consider models that utilize knowledge bases such as ConceptNet and Wikipedia, as well as models that utilize web search APIs to obtain external knowledge. \modelname{} incorporates COMET, which is trained on ConceptNet and ATOMIC, and we compare favourably against these models. Notably, \modelname{} achieves an accuracy of 43.14 on OK-VQA, outperforming KRISP (Wikipedia + ConceptNet + VQA P.T.) by over 4 points, and MAVEx (Wikipedia + ConceptNet + Google Images) by about 2 points. While our model clearly outperforms previous methods that use knowledge bases, it does not outperform models with large-scale pre-training and large number of parameters such as GPT-3 \cite{gpt3} and GPV2 \cite{gpv2}, which incorporate implicit commmonsense knowledge and require extensive resources to train. However, on OK-VQA, we achieve very similar results to PICa-Base \cite{yang2021empirical}, despite not having access to GPT-3. We expect that the use of a large pre-trained model like GPT-3 can further boost the performance of \modelname{}.

\subsection{Ablation Tests}
\label{sec:eval:ablations}
We perform comprehensive ablations on the validation set of the A-OKVQA dataset, as represented in Table~\ref{table:ablation}.\footnote{We present additional ablations in supplementary material Sec 2.3}
\paragraph{VQA P.T} We begin by training A-OKVQA on the baseline VL-BERT model without VQA pre-training. This gives us a score of 36.24. Next, obtain a new baseline for our model with VQA pre-training, where we then initialize \modelname{} with pre-trained weights on the VQA 2.0 dataset, and further train it on the A-OKVQA dataset. This results in a score of 43.46, over 7 points better, highlighting the impact of pre-training with a large-scale dataset. This model is a strong baseline for our VQA tasks.
\paragraph{Comm. Inference Representation} In the full model, we use SBERT to summarize each commonsense inference into a single vector, and use the multi-head attention block to capture useful information from the list of inference vectors. To test the effectiveness of our commonsense inference representation method, we first ablate SBERT, i.e., we incorporate all inferences as an additional text input for \modelname{}, feeding them token-by-token. This results in an accuracy score of 43.44, which is slightly lower than our baseline with VQA pre-training. Next, we use SBERT to summarize inferences, and feed the SBERT embeddings directly into \modelname{} with only a linear projection layer rather than the $\operatorname{MHA}$ block. This variant performs worse than the model with the $\operatorname{MHA}$ block by 1.25 points.
\paragraph{Augmented SBERT} In order to familiarize SBERT with our question-inference pairs, we fine-tune SBERT on the training set of A-OKVQA and OK-VQA  (Sec \ref{sec:method:comet:knowledge_selection}). We perform an ablation by evaluating our model on SBERT that has never been exposed to the question-inference-pairs. This results in a drop of 0.85 points in accuracy, which shows that our augmentation of SBERT is effective.

\section{Analysis}
\label{sec:analysis}
\subsection{Commonsense subsets}

\begin{figure*}[t]
\centering
    \includegraphics[width=0.9\textwidth, scale=0.1]{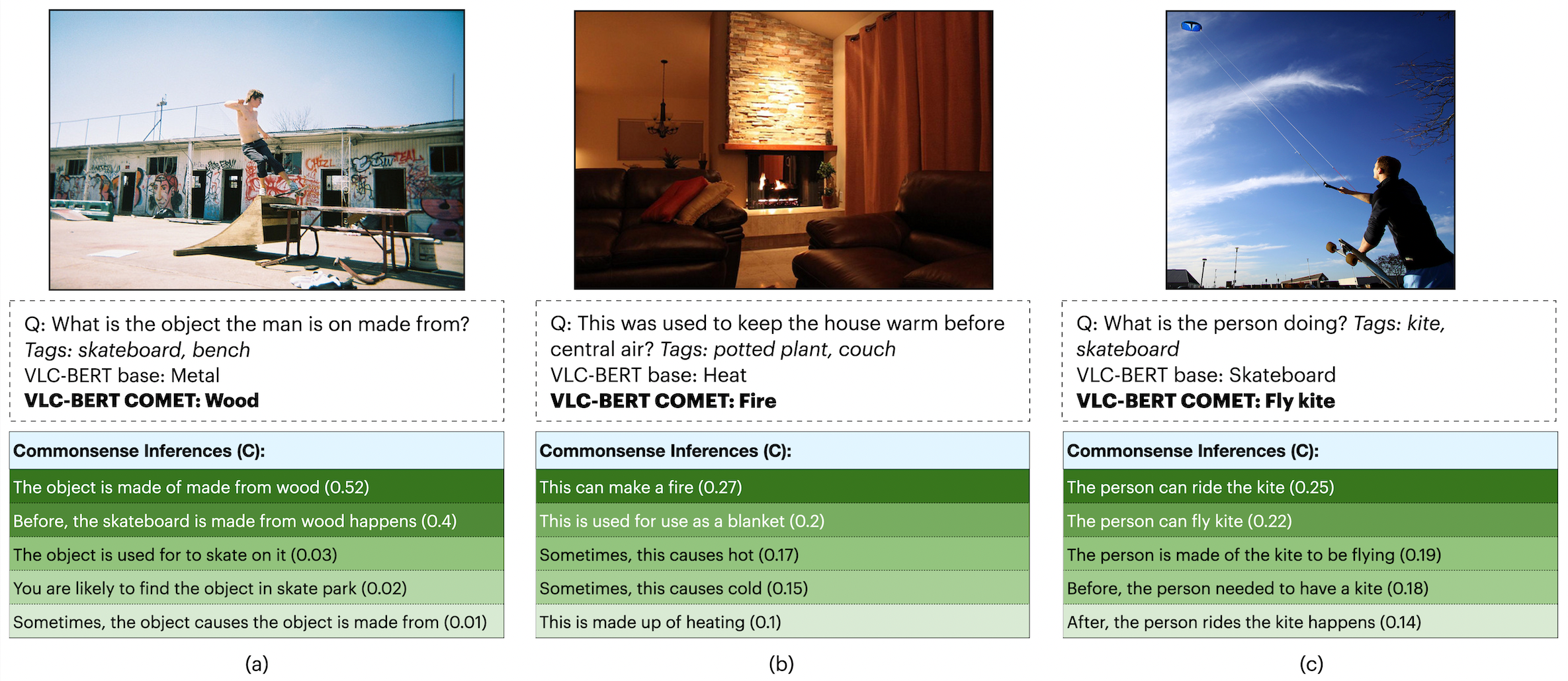}
    \vspace{-5pt}
    \caption{\textbf{Attention analysis:} (a) is from A-OKVQA, and (b) and (c) are from OK-VQA. We observe that the weakly supervised attention layer in \modelname{} accurately picks useful commonsense inferences. In (c), we observe how object tags are useful to guide COMET to produce contextualized knowledge.}
    \label{fig:egs}
\end{figure*}

\setlength{\tabcolsep}{4pt}
\begin{table}[t]
\begin{center}
\caption{Evaluation on the subsets of OK-VQA test (OK$_{s}$) and A-OKVQA validation (A-OK$_{s}$) sets, where factual, numerical and visual questions are pruned. The performance gain observed on the subsets shows a better picture of where external commonsense is effective.}
\label{table:subset}
\begin{tabular}{lcccc}
\hline\noalign{\smallskip}
Method & OK & OK$_{s}$ & A-OK & A-OK$_{s}$ \\
\noalign{\smallskip}
\hline
\noalign{\smallskip}
Base & 42.29 & 47.4 & 43.46 & 46.52 \\
w/ COMET & \textbf{43.14} & \textbf{48.21} & \textbf{44.95} & \textbf{49.53} \\
\hline
\end{tabular}
\end{center}
\end{table}
\setlength{\tabcolsep}{1.4pt}
\label{sec:datasets:eval_subsets}
Questions in OK-VQA and A-OKVQA datasets are diverse and require commonsense reasoning, visual understanding, as well as factual knowledge. While COMET can generate contextualized commonsense knowledge, it does not help with questions that require scene understanding (\eg, \quotes{What is to the left of the computer?}), factual knowledge (\eg, \quotes{Where was this food invented?}), or text/symbol recognition (\eg, \quotes{What does this sign say?}). Moreover, averaging results on the entirety of OK-VQA and A-OKVQA obfuscates the improvements brought about to a subset of questions that truly require commonsense knowledge. We propose subsets to assess the performance of our model on questions that are more likely to require external commonsense knowledge. We obtain the subsets by eliminating questions that are mostly factual or visual, and hence do not require commonsense, following these conditions: (1) \emph{factual}: The question or answer contains named entities (\eg, \quotes{USA}); (2) \emph{numerical}: The answers contain numbers or number words (\eg, \quotes{twenty}) or the question has date or time words (\eg, \quotes{century}); (3) \emph{visual}: The question contains directional words (\eg, \quotes{left of}) and words referring to symbols (\eg, \quotes{mascot}).

In Table \ref{table:subset}, we show that \modelname{} with COMET performs 3 points better on the A-OKVQA subset, and maintains an 0.8 point improvement on the OK-VQA subset. This substantiates our claim that utilizing our COMET pipeline substantially increases \modelname{}'s ability to answer questions that require external knowledge. %However, it also highlights the need for ground truth question-type annotations.

\subsection{Attention Analysis} 
In this section, we show qualitative examples to demonstrate questions where \modelname{} benefits from contextualized commonsense knowledge from COMET. We also show the corresponding attention weights, to show the effectiveness of the proposed weakly-supervised attention mechanism. Fig~\ref{fig:egs}a shows an example from A-OKVQA, where COMET's inferences on the question and the object tags, weighted by the attention score, results in the correct answer. Fig~\ref{fig:egs}b shows an example from OK-VQA where \modelname{} COMET exhibits higher attention towards the fire despite the object tags missing the fireplace. This is an example where deriving inferences from the question phrase is equally important as doing so with the object tags. Fig~\ref{fig:egs}c shows that inferences on the object tag \textit{kite} drove the model to answer correctly. The supplementary material includes additional examples of improvements and failures. 

% \section{Discussion}
% \label{sec:discussion}
% \input{sections/8-discussion}

\section{Conclusions}
\label{sec:conclusions}
We presented Vision-Language-Commonsense BERT (\modelname{}) for external knowledge-driven VQA tasks. \modelname{} outperforms previous models based on knowledge bases on the OK-VQA and A-OKVQA datasets by incorporating contextualized commonsense knowledge from COMET and combining it with visual and linguistic inputs. Through our evaluation, we show the effectiveness of our knowledge generation, selection, and incorporation strategies, and the positive impact of VQA pre-training.

Our analysis of \modelname{} highlighted a few limitations of our model and the datasets we evaluate on. First, some questions require a deeper understanding and linking of multiple entities and events in the image, that object tags lack, for deriving relevant commonsense inferences. Second, condensing the commonsense inferences using SBERT and $\operatorname{MHA}$ leads to a compressed representation that may cause the model to lose some information. Finally, our model is limited by COMET, and the knowledge bases it is trained on, as we observe that large-scale models like GPT-3 outperform it. 

 We view our work as a first step in analyzing the potential of \textit{generative commonsense incorporation}, and exploring approaches to \textit{decide when commonsense is needed}. In the future, our goal is to work towards creating a version of COMET that can utilize image context concerning multiple entities and events. We also plan to investigate the potential of multi-hop reasoning with COMET to bridge the question and image-based expansions closer.

\section{Acknowledgments}
\label{sec:acknowledgments}
This work was funded, in part, by the Vector Institute for AI, Canada CIFAR AI Chair, NSERC CRC, NSERC DG and Accelerator Grants, and a research gift from AI2. Hardware resources used in preparing this research were provided, in part, by the Province of Ontario, the Government of Canada through CIFAR, and companies sponsoring the Vector Institute\footnote{\href{www.vectorinstitute.ai/\#partners}{www.vectorinstitute.ai/\#partners}}. Additional hardware support was provided by John R. Evans Leaders Fund CFI grant and Compute Canada under the Resource Allocation Competition award.
%Additional support was provided by JELF CFI grant and Compute Canada under the RAC award.
Finally, we sincerely thank Prof. Giuseppe Carenini for valuable feedback and discussions.

{\small
\bibliographystyle{ieee_fullname}
\bibliography{main}
}

\end{document}